\documentclass[letterpaper, 10 pt, conference]{ieeeconf}  
\usepackage{graphicx}
\usepackage{amsmath} 
\usepackage{booktabs}
\usepackage{multirow}
\usepackage{placeins}

\IEEEoverridecommandlockouts                              

\title{\LARGE \bf
UniTac-NV: A Unified Tactile Representation For Non-Vision-Based Tactile Sensors*
}

\author{Jian Hou$^{1}$,~\IEEEmembership{Student Member,~IEEE}, Xin Zhou$^{1}$,~\IEEEmembership{Student Member,~IEEE}, Qihan Yang$^{1}$,~\IEEEmembership{Student Member,~IEEE} \\ and Adam J. Spiers,~\IEEEmembership{Member,~IEEE}
\thanks{*This research was supported by The Royal Society (Grant No. RGS/R2/222382) and Imperial College London internal funds.}
\thanks{All authors are with the Manipulation and Touch Lab, Dept. Electrical
and Electronic Engineering, Imperial College London. (Corresponding Author: jian.hou22@imperial.ac.uk)}
\thanks{$^{1}$These authors contributed equally to this work.}
}

\begin{document}

\maketitle
\thispagestyle{empty}
\pagestyle{empty}

\begin{abstract}

Generalizable algorithms for tactile sensing remain underexplored, primarily due to the diversity of sensor modalities. Recently, many methods for cross-sensor transfer between optical (vision-based) tactile sensors have been investigated, yet little work focus on non-optical tactile sensors. To address this gap, we propose an encoder-decoder architecture to unify tactile data across non-vision-based sensors. By leveraging sensor-specific encoders, the framework creates a latent space that is sensor-agnostic, enabling cross-sensor data transfer with low errors and direct use in downstream applications. We leverage this network to unify tactile data from two commercial tactile sensors: the Xela uSkin uSPa 46 and the Contactile PapillArray. Both were mounted on a UR5e robotic arm, performing force-controlled pressing sequences against distinct object shapes (circular, square, and hexagonal prisms) and two materials (rigid PLA and flexible TPU). Another more complex unseen object was also included to investigate the model's generalization capabilities. We show that alignment in latent space can be implicitly learned from joint autoencoder training with matching contacts collected via different sensors. We further demonstrate the practical utility of our approach through contact geometry estimation, where downstream models trained on one sensor's latent representation can be directly applied to another without retraining. 

\end{abstract}

\section{INTRODUCTION}

Tactile sensing can enhance robotic perception, especially in scenarios involving visual occlusions or limited lighting. In addition, some object properties are inherently challenging to infer via vision, such as detailed surface shape, friction, stiffness and texture. Researchers have demonstrated that a wide range of use cases benefit from tactile sensing, including object pose estimation/contact characterization during in-hand manipulation (where fingers frequently occlude objects) \cite{yin2023rotating, saloutos2023design, hou2024location}, object identification and characterization \cite{bottcher2021object, zhou2023tactile, spiers2016single, zhou2022troll}, and grasp stability estimation \cite{saloutos2023towards, grover2022learning, khamis2018papillarray}.

Despite their importance, tactile algorithms and methods are not widely available, especially compared to the accessibility of computer vision methods. This is in part due to the wide range of sensing mechanisms and modalities present in current available tactile sensors, making generalizable approaches rare. For example, the DIGIT sensor \cite{lambeta2020digit} produces high-resolution RGB images capturing sensing surface deformations, whilst the Contactile PapillArray \cite{khamis2018papillarray} produces 3-DoF force-torque readings from each distinct pea-sized sensing unit laid out in a matrix.
Another challenge is sensor obsolescence: the once relatively popular BioTac \cite{fishel2012sensing} and TakkTile \cite{takk1} sensors are now obsolete, with no comparable alternatives available. As a result, significant portions of research works involving these sensors are now challenging to continue or have lost relevance.

\begin{figure}[t]
    \centering
    \includegraphics[width=1\linewidth]{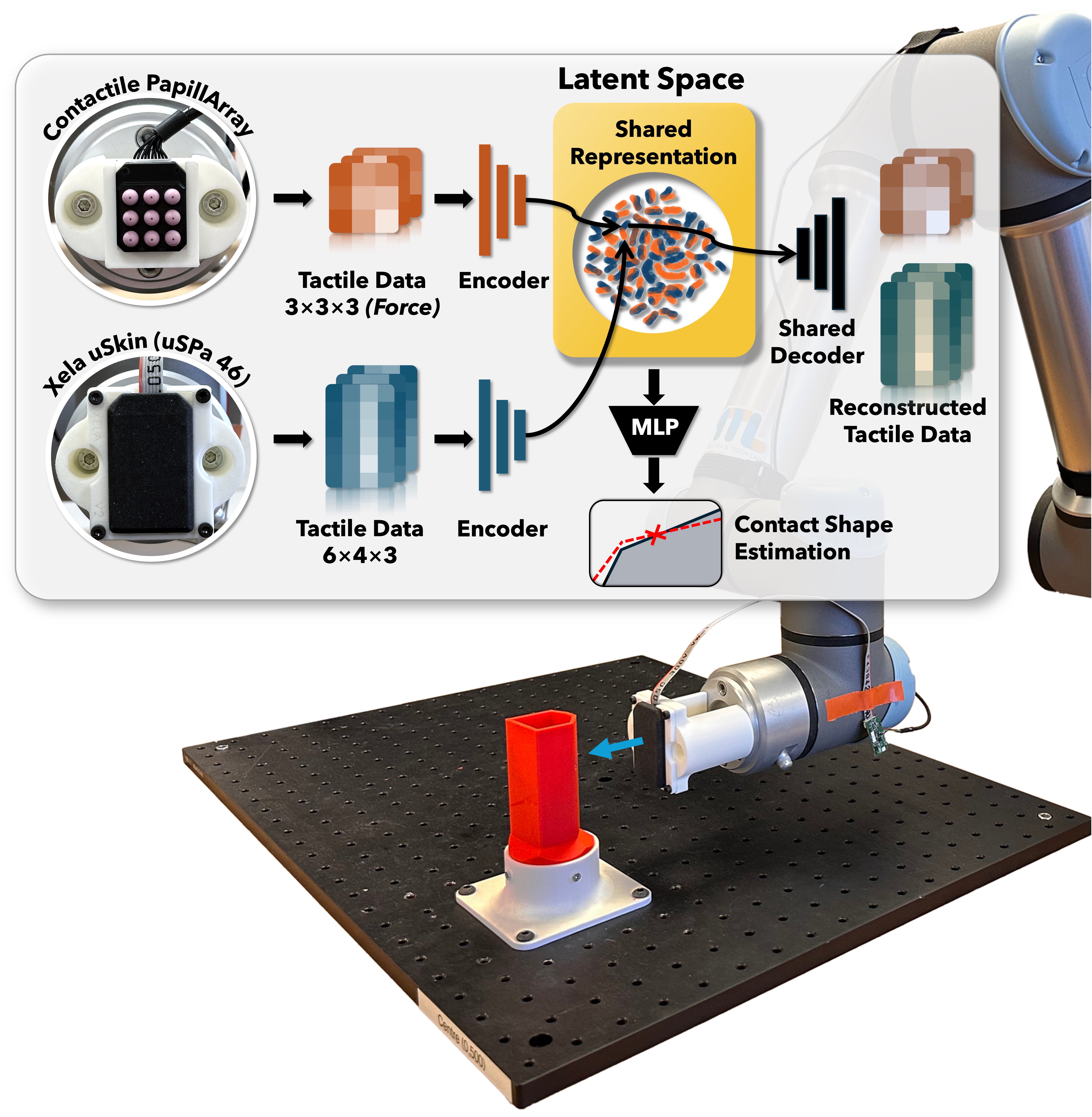}
    \caption{The UniTac-NV autoencoder architecture, which unifies tactile data from two different sensors into a shared latent space via training on matching tactile interactions collected on different sensors. This unified representation provides two main use cases: 1. The shared decoder can reconstruct tactile information from the latent space, enabling cross-sensor data transfer. 2. Downstream models trained on one sensor's latent representation can be directly applied to another without retraining.}
    \label{system}
\end{figure}

Vision-based tactile sensors represent a special case, with examples such as the aforementioned DIGIT \cite{lambeta2020digit}, GelSight \cite{yuan2017gelsight}, and TacTip \cite{ward2018tactip}. These sensors use cameras to capture micrometer-level deformations of an elastomer sensing surface, ideal for recognizing complex and detailed contact shape features as demonstrated in \cite{potdar2024high}. They also share similar internal structures and produce image outputs of comparable formats, which led to applications and methods often being generalizable among them. Numerous recent works have demonstrated that a unified representation across different vision-based tactile sensors can be learned, which enables data collected on one sensor model to be shared across others. For example, \cite{higuera2024sparsh} pretrained a self-supervised model on several large-scale datasets of vision-based tactile data, including Digit, GelSight, and GelSight Mini, to create a unified representation transferrable to each sensor. Another approach aligns tactile data from two different sensors via contrastive learning \cite{rodriguez2024contrastive}. In \cite{zhao2024transferable}, task-specific decoders have been used to train individual encoders, even with large unpaired datasets.

However, vision-based tactile sensors are not a one-size-fits-all solution. Their limited operating frequencies (often 30 Hz) and high latencies (between 100-200 ms) make them unideal for fast dynamic tasks, whilst their bulky form factors (often \> 40 mm) make them challenging to fully cover robotic hands \cite{shah2021design}. In such cases, other non-vision-based tactile sensors are more suitable: 
the Xela uSkin provides various shapes and taxel configurations (ranging from 1 $\times$ 1 to 4 $\times$ 6 sensing units) with a thin profile (around 5 mm) and can cover large portions of robotic hands such as the Allegro hand \cite{chelly2024interaction}; the Contactile PapillArray is better at measuring dynamic 3-DoF force and at high temporal resolution (1000 Hz maximum), enabling tasks such as real-time slip detection \cite{khamis2018papillarray}. However, little work has focused on representation learning and cross-sensor information sharing for these non-vision-based sensors.  

To address this gap, we introduce the Cross-Sensor Tactile  Representation Autoencoder (UniTac-NV), an encoder-decoder network featuring dedicated encoders for each sensor and a shared decoder for reconstructing tactile sensor readings (Fig. \ref{system}). Our approach encodes contact information into a unified latent space that is agnostic to the sensor type. The shared decoder then translates this latent representation into the raw readings for all tactile sensors involved.

The advantages of our architecture are two-fold: Firstly, we demonstrate that the shared latent space can be used directly in application tasks with minimal loss of information. Conceptually, one can view the decoders as translators that convert sensor-specific data (raw sensor readings) into a universal `common language' represented by the latent space. Consequently, downstream applications built on this common representation can seamlessly integrate any tactile sensor whose data can be translated into this universal format. Secondly, our loss function—derived from tactile reconstruction errors—implicitly aligns sensor-specific encodings, as demonstrated later. This streamlined approach simultaneously trains the encoders and the shared decoder, naturally supporting cross-sensor translation---an added benefit compared to methods that focus solely on latent alignment.

We validate our approach using two commercial non-vision-based tactile sensors: the Xela uSkin Patch (Model: uSPa 46) and the Contactile PapillArray.
Our experiments quantify both cross-sensor translation accuracy and latent space alignment. We further assess the framework by performing a downstream task: contact geometry reconstruction.
In addition, we propose a reproducible tactile data collection procedure which ensures consistent contact interactions across different tactile sensors. This allows us, as well as the tactile sensing community, to expand the dataset with new contact interactions as well as sensors.

\section{METHODOLOGY}

\subsection{The UniTac-NV Autoencoder Model}
\subsubsection{Multi-Encoder Architecture}
UniTac-NV (shown in Fig. \ref{system}) was inspired by traditional autoencoder architectures, which learn compact latent representations by minimizing errors between original and reconstructed input data. In contrast to traditional autoencoders, UniTac-NV features multiple encoders (one for each tactile sensor) and one shared decoder, which reconstructs tactile readings from the latent space for all involved sensors. Each encoder takes the flattened sensor data as input (4 $\times$ 6 $\times$ 3 = 72 readings for uSkin and 3 $\times$ 3 $\times$3 = 27 force readings for PapillArray) and consists of two hidden dense layers (64 and 48 units), followed by a projection to a 16-dimensional latent space. The shared decoder processes the latent data through two hidden dense layers (64 and 96 units) and outputs a 72 + 27 = 99-dimensional reconstruction, combining the uSkin's and PapillArray's readings. In addition, dropout layers are put between all layers. 

\subsubsection{Sample-Matched Training} \label{trainingStrategy}
Our training strategy is key to aligning latent spaces across sensors. In each training pass, we feed matching contact interactions (same contact shapes and pressing forces) from the two different sensors into the corresponding sensor encoders. The encoders output two latent representations in total, which the decoder then uses to generate four reconstructions: two self-reconstructions (each sensor's latent representation reconstructing its own input) and two cross-reconstructions (each latent representation reconstructing the input from the other sensor). The final loss is computed as the sum of the four reconstruction errors:
\[
\mathcal{L}_{\text{total}} = \sum_{\substack{i,j \in \{1,2\}}} \text{MAE}\big(X_j, \hat{X}_j^{i}\big) \quad  \label{eq:total_loss},
\]
where $X_j$ is the raw input data from sensor \(j\) and $\hat{X}_j^{i}$ is the reconstruction of sensor  \(j\)'s data from sensor \(i\).



This strategy helps the latent space capture sensor-agnostic physical properties rather than noise. Alignment occurs because the decoder requires consistent latent features to accurately reconstruct the paired input data.

\subsubsection{Single-Encoder Inference}\label{inferenceStrategy}

After model training, we have jointly optimized one encoder per tactile sensor as well as a shared decoder capable of translating the shared latent representation into both tactile sensor formats. One can then either select the corresponding encoder to translate tactile data into the common latent representation directly for downstream tasks, or transfer the data to a different tactile sensor format.


    

\subsection{Hardware Setup}

\begin{figure}[t]
    \centering
    \vspace{6pt}
    \includegraphics[width=1\linewidth]{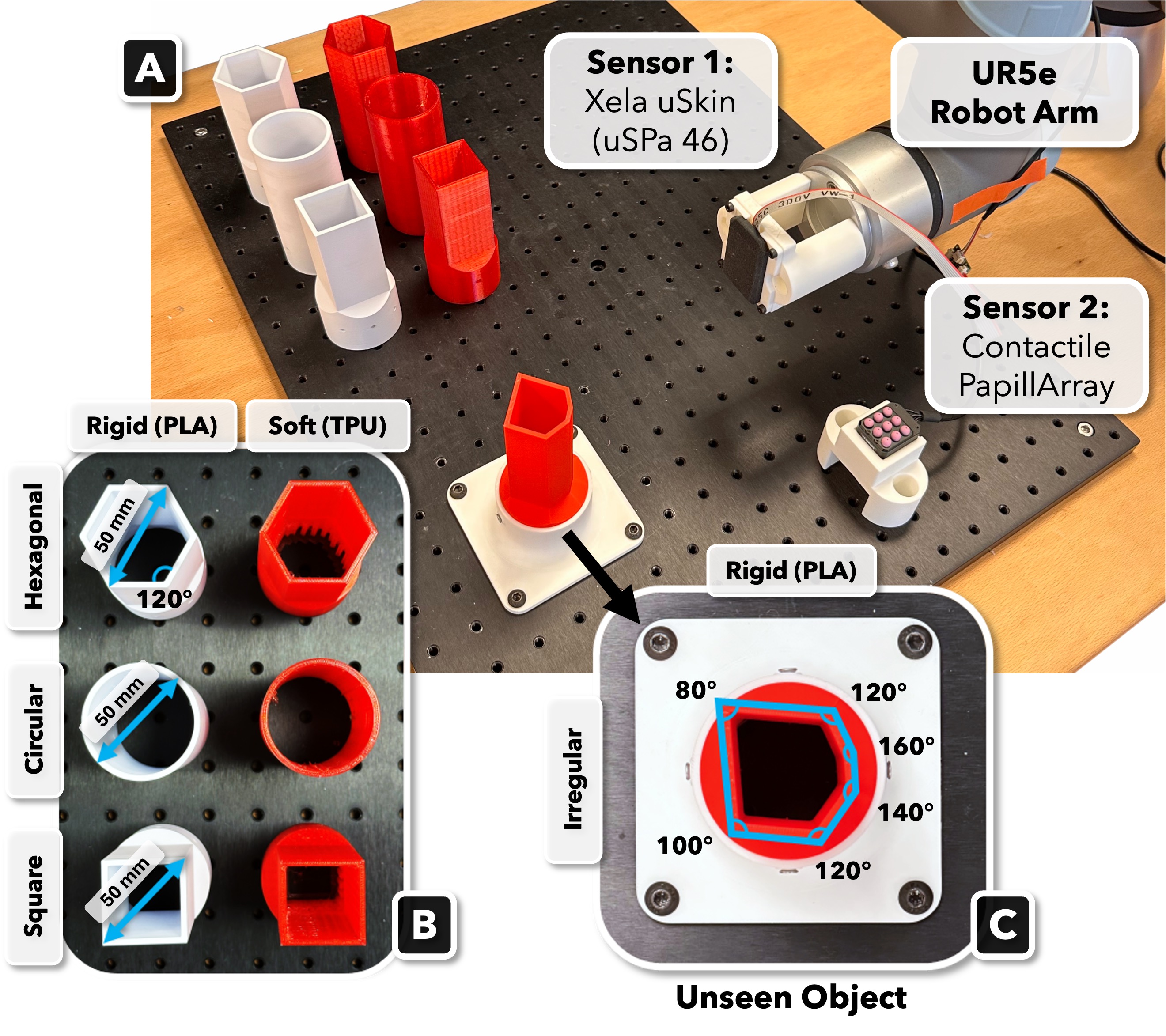}
    \caption{Hardware setup for tactile contact data collection. A: The UR5e robotic arm with two exchangeable tactile sensors—the XELA uSkin (uSPa 46) and the Contactile PapillArray. B: Three different shapes (square, circular, and hexagonal prisms) are 3D-printed in two materials each (rigid PLA and soft TPU). C: An irregular unseen object, 3D-printed in PLA, consisting of six corners with angles ranging from 80 to 160 degrees.}
    \label{setup}
\end{figure}

Fig. \ref{setup} shows the hardware setup for experiments. A UR5e robotic arm with Robotiq Force Copilot software performs consistent and force-monitored movements for data collection (outlined in section \ref{Section_DataCollection}). We use a Thorlabs MB1224 aluminium optical breadboard to ensure experiment objects are secured at precisely known positions via a 3D printed PLA mounting adapter (white) shown in Fig.\ref{setup}(B). Objects are fastened to this adapter via four horizontal screws, which also allow objects to be rotated in precise 90-degree steps.

Two tactile sensors have been used in this work, both configured to operate at 100 Hz: 
\begin{itemize}
    \item XELA uSkin (uSPa 46): A Hall-effect-based tactile sensor with a sensing area around 30 mm $\times$ 50 mm and 4$\times$6 sensing units, each detecting 3-DoF force. The sensing units are covered by a thin sheet of elastic material to create a continuous sensing surface.
    \item Contactile PapillArray: A tactile sensor with a sensing area around 24 mm $\times$ 24 mm featuring 3-DoF optical force sensing units. In contrast to the uSkin, the PapillArray only has 3$\times$3 sensing units which are directly exposed instead of under a continuous `skin'. The center sensing unit is also slightly more protruded compared to the other eight, as opposed to the flat sensing area of the uSkin.   
    
\end{itemize}

Two customized sensor adapters attach the sensors to the UR5e’s tool flange. Their dimensions ensure that the two sensors' contact surface centers are in the same pose relative to the tool flange, ensuring data collection consistency.

\subsection{Experiment Objects}

Fig. \ref{setup}(B) and Fig. \ref{setup}(C) list the experiment objects. A set of 6 `seen' objects consists of hollow prisms with circular, square, and hexagonal bases. Each shape has two material versions: one rigid version, 3D-printed in PLA material on a Bambulab P1S printer, and another soft version printed with TPU material on a Prusa MK3S. The hollowness ensures that the soft versions deform upon the force applied. A more irregular `unseen' rigid PLA prismatic object is included for testing, featuring corners ranging from 80 to 160 degrees with 20-degree steps.

\subsection{Data Collection}\label{Section_DataCollection}

\begin{figure}[t]
    \centering
    \vspace{6pt}
    \includegraphics[width=1\linewidth]{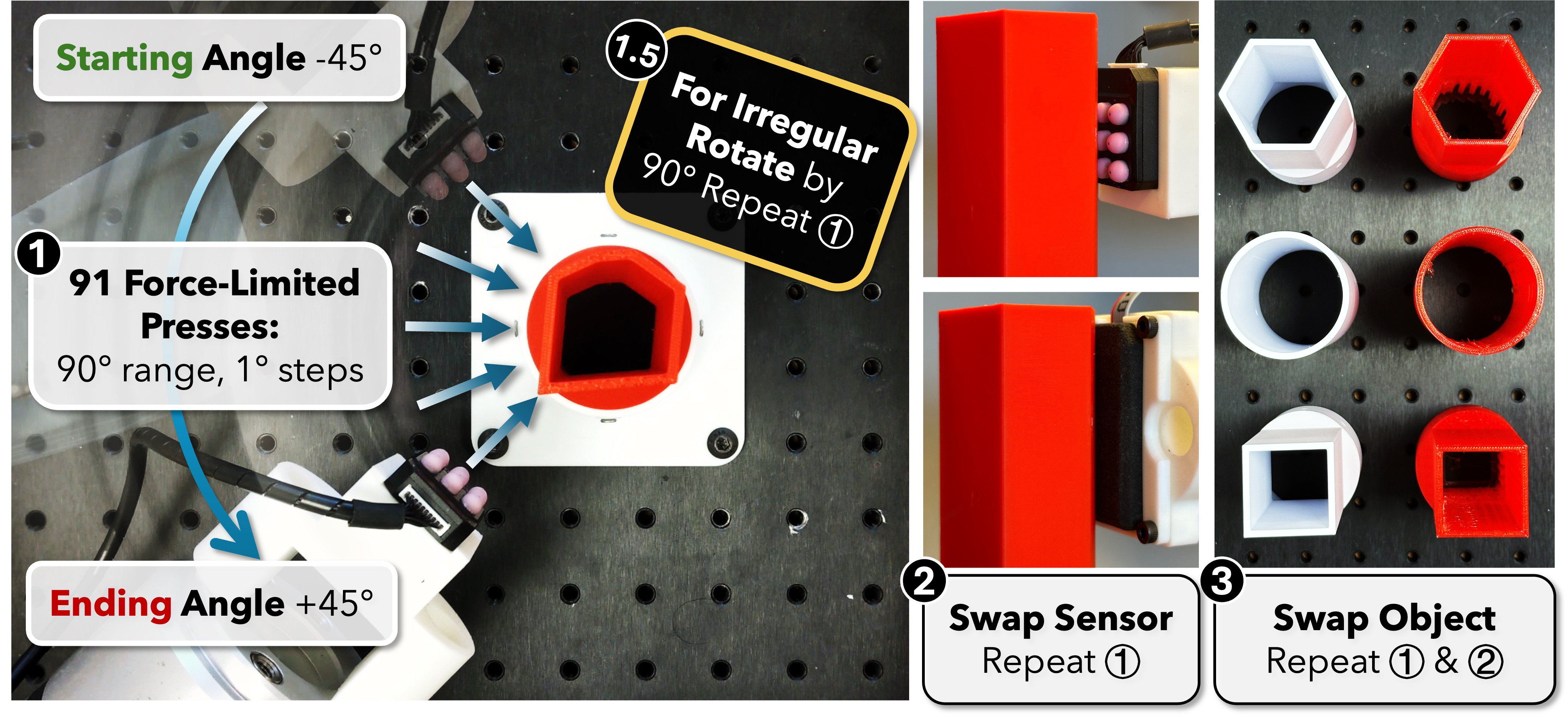}
    \caption{Data collection procedure. Step 1: Mount the object on the optical breadboard and attach the first sensor to the arm's tool flange. Perform force-controlled presses along a 90-degree arc towards the object's center at a constant height. This step is performed 4 times on the irregular object, with 90-degree object rotations in between, to ensure full coverage of its outline. Step 2: Interchange the sensors and repeat Step 1 to collect matching tactile data across sensors. Step 3: Repeat the previous steps for all objects involved. }
    \label{data_collection}
\end{figure}
Fig. \ref{data_collection} outlines our proposed data collection procedure. Essentially, we repeatedly press the sensors against different objects at different angles whilst ensuring the pressing positions, angles, and forces are consistent between the two tactile sensors for each object. Each press is performed horizontally at a chosen height, targeting the center-line of the object adapter. We utilize Robotiq's Force Copilot software to accurately monitor pressing forces from the UR5e's force-torque sensor and terminate each press once 10 N of z-axis force is reached. Since most objects in this experiment are symmetric, we limit our range of pressing angles to 90 degrees for efficiency and perform 91 presses with 1-degree steps between consecutive presses. For each object, this same procedure is performed for both sensors, resulting in two time-series tactile data with monitored UR5e tooltip forces. We then take 25 target tooltip's z-axis force from 4 to 10 Newtons with 0.25-Newton intervals, and extract tactile data at the closest-matching time frames for each sensor, approach angle, and object. This finally results in a dataset of tactile data that have matching 2,275 contact interactions between the two sensors for each object.









\section{UniTac-NV Training \& Evaluation}
\label{sec3}
\subsection{Data Preparation \& Model Training} \label{section_dataPrep}
\begin{figure}[t]
    \centering
    \vspace{3pt}
    \includegraphics[width=1\linewidth]{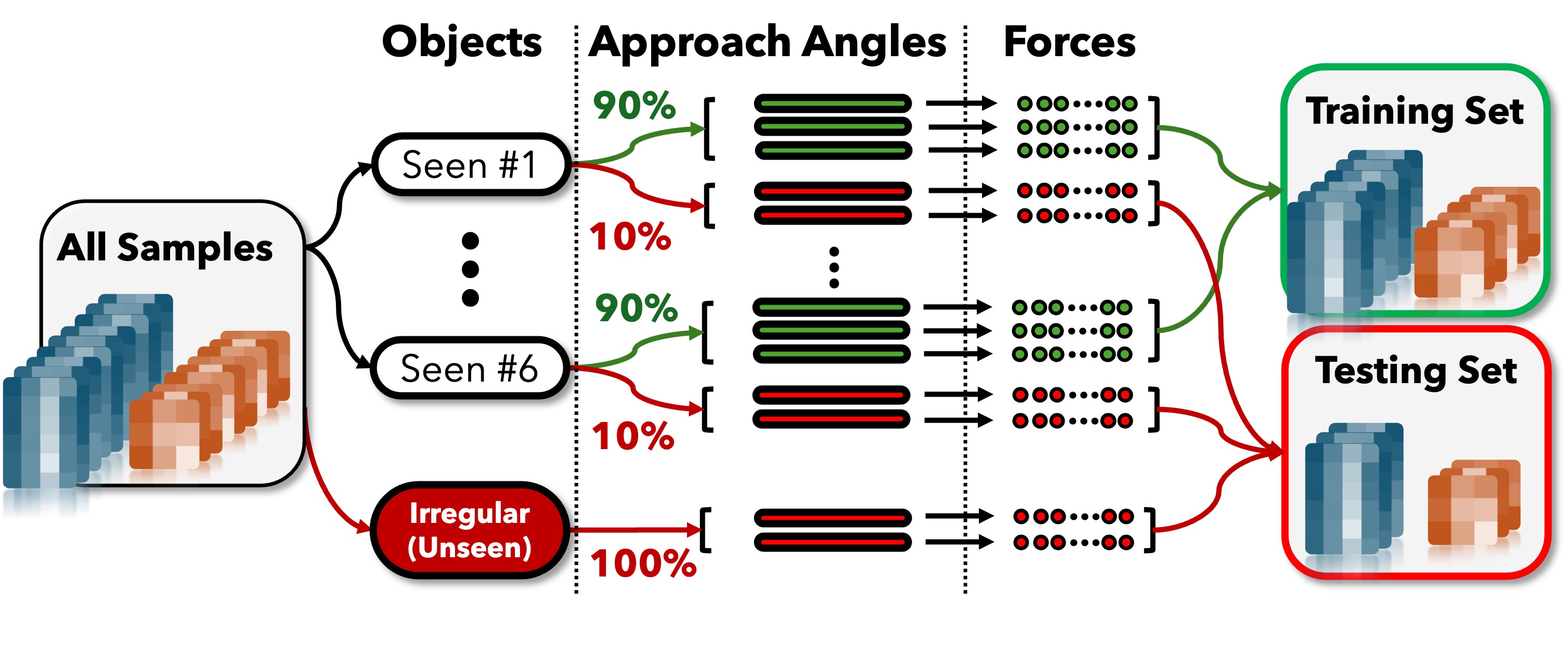}
    \caption{Overview of the data splitting strategy. Matched between sensors, 10\% of the approaching angles are randomly selected for each object, and their corresponding data samples (varying in pressing forces) are included in the testing set. This approach prevents data leakage while ensuring paired samples that are consistent in the object, approaching angle, and force. All data collected from the irregular unseen object are added to the testing set.}
    \label{trainTest}
\end{figure}

Collected data from the six seen objects are split into 90\% training and 10\% testing data. We follow a stratified approach (outlined in Fig. \ref{trainTest}) to create balanced yet randomized data sets: First, data splitting is performed on each object separately, resulting in each object being equally represented in both training and testing sets. Secondly, for each object, we pseudo-randomly select 10\% of all approaching angles and include all contained data samples (varying in pressing forces) in the test set. This prevents data leakage by ensuring that nearly identical tactile samples—collected from the same object and approaching angle with slight force variations—do not cross data sets. Finally, 100\% of the data collected on the irregular unseen object are added to the testing set. Note that following the training strategy outlined in section \ref{trainingStrategy}, the same data preparation steps are followed for both tactile sensors, resulting in paired samples with matching contact interactions. Following the training strategy in section \ref{trainingStrategy}, we trained the model with 1000 epochs on the training set with the Adam optimizer \cite{kingma2014adam}, a 5e-4 learning rate, a dropout rate of 0.007, and a batch size of 64.

After training our UniTac-NV model on the training set, we assess its performance via the inference method described in section \ref{inferenceStrategy}: each tactile reading sample (from either sensor) is fed into their corresponding encoder. The decoder then produces two outputs---self-reconstructed tactile data and tactile data transferred to the other sensor (cross-reconstructed). The self-reconstruction quality reflects how well the tactile readings can be recovered from the latent space, and cross-reconstruction error measures the performance of transferring tactile data between sensors.

\subsection{Metrics} \label{metrics}
\begin{figure}[!b]
    \centering
    \includegraphics[width=1\linewidth]{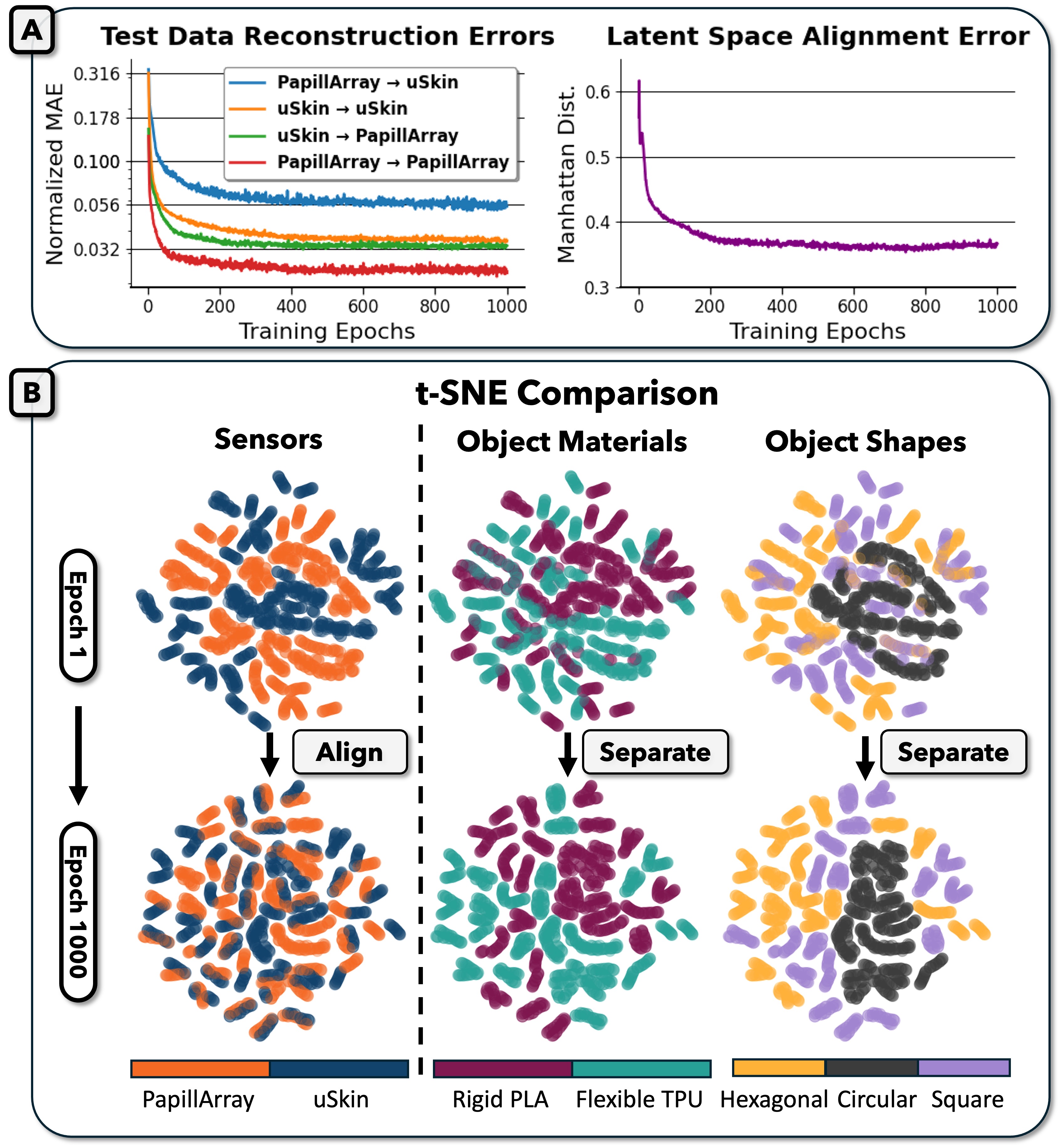}
    \caption{
    A: Testing samples' reconstruction errors (Normalized MAEs) and the latent space alignment error over the 1000 epochs. The reconstruction errors include both self-reconstruction (same sensor) and cross-reconstruction (between sensors). B: t-SNE visualizations of latent space distributions after the first and final training epochs. Tactile samples matching in contact interaction but varying in sensor type are implicitly aligned through training, while samples varying in object material and object shape are separated.}
    \label{latent}
\end{figure}
We want to test the framework's ability to reconstruct tactile data from the latent representation and demonstrate that the latent encodings from different sensors are aligned but separable in object shape and material (can be used as a unified common representation with low information loss).

To quantify the data reconstruction errors, two metrics were implemented to measure the discrepancies between the original ($X$) and reconstructed ($\hat{X}$) sensor data. We calculated the normalized mean absolute error (NMAE) and the structural similarity index (SSIM) between them across all sensing units and three force directions (x, y, z). SSIM is better at capturing the discrepancies in spatial force distribution and relative magnitudes~\cite{wang2004image} and calculated as:



\begin{equation}
    \text{SSIM}(X, \hat{X}) = \frac{1}{3} \sum_{c \in \{x,y,z\}} \text{SSIM}_c(X^c, \hat{X}^c) \notag
\end{equation}

\[
\begin{aligned}
&\text{SSIM}_c(X^c,\hat{X}^c) \\
&= \frac{(2\mu_{X}^c\,\mu_{\hat{X}}^c + C_1)(2\sigma_{X\hat{X}}^c + C_2)}{((\mu_X^c)^2+(\mu_{\hat{X}}^c)^2+C_1)((\sigma_X^c)^2+(\sigma_{\hat{X}}^c)^2+C_2)}, 
\end{aligned}
\]
where $c$ is the tactile readings' force direction channel ($x$, $y$, and $z$), and $\text{SSIM}_c$ is the per force-channel $\text{SSIM}$. $\mu$ is the mean across a force channel, $\sigma$ the standard deviation, and $C_1$ and $C_2$ are relatively small constants that help stabilize the division. 

Manhattan distance is used to quantify alignment quality between latent spaces generated from each sensor's encoder.

\subsection{Implicit Latent Space Alignment \& Separation}
\begin{figure*}[t]
    \centering
    \vspace{8pt}
    \includegraphics[width=\linewidth]{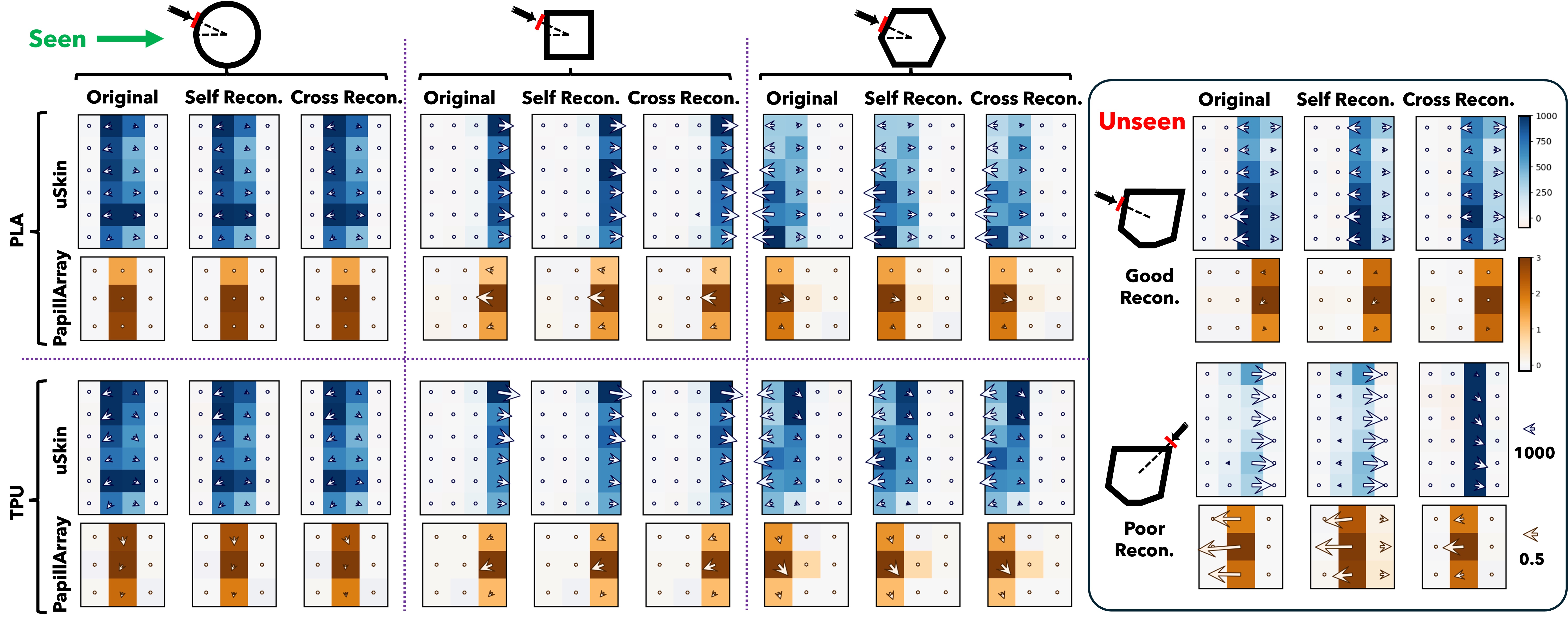}
    \caption{Examples of tactile data reconstructions for both seen and unseen objects. Each cell represents a sensing unit, with color indicating the normal force and arrows showing shear forces. Reconstructions include both self-reconstruction (same sensor) and cross-reconstruction (different sensors). We selected the samples from the same pressing angle for seen objects. For unseen objects, two samples were chosen: one with good reconstruction and another with relatively poor reconstruction. Poor reconstructions for unseen objects often occur when the pressing involves a sharp edge with a small contact area.}
    \label{reconstruction}
\end{figure*}

Fig. \ref{latent}(A) presents the reconstruction errors and latent space distances between matched data samples from different sensors during 1000 training epochs. Both metrics were calculated on the testing set, which was not involved in training. Intuitively, cross-sensor reconstructions performed slightly worse on average than same-sensor reconstructions. This is particularly clear when transferring PapillArray data to uSkin data, the uSkin features nearly three times as many sensing units requiring significant upsampling. Fig. \ref{latent}(B) compares the latent spaces encoded from the two sensors after the first and the final training epoch. Latent representations of all test samples are visualized in 2D via t-distributed stochastic neighbor embedding (t-SNE) \cite{van2008visualizing}. A successful alignment encoding should align the latent spaces of two different sensors, ensuring that their representations are unified while preserving the original information. From these t-SNE plots, it is clear that in addition to the latent spaces being much more aligned between two sensors, they also are much more separable in object material (rigid vs soft) and shape after training. This indicates that our sample-matched training method outlined in section \ref{trainingStrategy} effectively preserves material and shape characteristics in a sensor-agnostic manner. Note that neither latent alignment nor separation are utilized as training losses---they are implicitly learned.

\subsection{Data Reconstruction Results: Self-Reconstruction and Cross-Sensor Transfer}

Fig. \ref{reconstruction} shows a small selection of data reconstruction examples as quiver plots after training. Each individual cell represents a sensing unit, with its color indicating force in the z-direction (normal to the sensing surface) and the arrow starting its center showing the shear force (combination of x and y forces). Visually, most reconstructions look similar if not indistinguishable from the original. TABLE \ref{table:result} quantifies the reconstructions via the NMAE and SSIM (outlined in section \ref{metrics}), both calculated on the test set. We compare UniTac-NV's performance against autoencoders trained separately on each sensor (with the same encoder and decoder architecture as the UniTac-NV), which serve as `near-maximum' performance levels for benchmarking. Generally, on seen objects, our model's reconstructions have high performances, with excellent structural similarities (SSIM $>$ 0.95) and normalized MAE errors slightly elevated compared to using individually trained autoencoders. On the irregular object that contains more complex geometries, the same-sensor reconstructions (uSkin to uSkin, PapillArray to PapillArray) are still comparable to reconstructions via individual autoencoders, with very similar structural similarities. However, whilst transferring uSkin data to the PapillArray performs well (SSIM $>$ 0.85), the other way around performs relatively poorly. This is most likely due to the inherent differences in sensor design and taxel dimensions: we are essentially upsampling (3 × 3 × 3) PapillArray data to (4 × 6 × 3) uSkin data.



\begin{table*}[t]
\centering
\vspace{5pt}
\caption{Normalized MAE / SSIM for Data Reconstruction}
\begin{tabular}{@{}cccccccc@{}}
\toprule
 &  & \multicolumn{2}{c}{Individually Trained Autoencoder} & \multicolumn{4}{c}{\textbf{UniTac-NV} Tactile Data Transfer} \\ \midrule
 &  & \multirow{2}{*}{\begin{tabular}[c]{@{}c@{}}uSkin \\ Recon.\end{tabular}} & \multirow{2}{*}{\begin{tabular}[c]{@{}c@{}}PapillArray \\ Recon.\end{tabular}} & \multicolumn{2}{c}{From uSkin} & \multicolumn{2}{c}{From PapillArray} \\ \cmidrule(l){5-8} 
 &  &  &  & To uSkin & To Papill & To Papill & To uSkin \\ \midrule
\multirow{3}{*}{PLA} & Circular & 0.014 / 0.996 & 0.008 / 0.997 & 0.018 / 0.984 & 0.020 / 0.992 
& 0.010 / 0.995 & 0.031 / 0.973 
\\
 & Square & 0.032 / 0.981 & 0.020 / 0.993 & 0.038 / 0.981 & 0.044 / 0.958 
& 0.031 / 0.973 & 0.064 / 0.957 
\\
 & Hexagonal & 0.034 / 0.965 & 0.016 / 0.996 & 0.041 / 0.955 & 0.031 / 0.985 
& 0.027 / 0.990 & 0.050 / 0.944 
\\ \midrule
\multirow{3}{*}{TPU} & Circular & 0.020 / 0.992 & 0.009 / 0.997 & 0.029 / 0.983 & 0.021 / 0.991 
& 0.013 / 0.994 & 0.041 / 0.955 
\\
 & Square & 0.032 / 0.972 & 0.020 / 0.990 & 0.037 / 0.962 & 0.040 / 0.950 
& 0.027 / 0.989 & 0.059 / 0.939 
\\
 & Hexagonal & 0.030 / 0.986 & 0.018 / 0.996 & 0.042 / 0.981 & 0.040 / 0.974 
& 0.030 / 0.990 & 0.066 / 0.947 
\\ \midrule
\multicolumn{2}{c}{All Seen} & 0.027 / 0.982 & 0.015 / 0.995 & 0.034 / 0.974 & 0.033 / 0.975 
& 0.023 / 0.989 & 0.052 / 0.953 
\\ \midrule
\multicolumn{2}{c}{Irregular (Unseen)} & 0.111 / 0.862 & 0.036 / 0.982 & 0.125 / 0.845 & 0.136 / 0.858 & 0.058 / 0.961 & 0.318 / 0.581 
\\ \bottomrule
\end{tabular}
\label{table:result}
\end{table*}

\section{Downstream Task: Contact Geometry Estimation}
In the previous section, we have trained the sensor-specific encoders as well as the shared decoder and have shown that: 1. self-reconstruction and cross-sensor translation have relatively low errors; 2. latent spaces encoded from different sensors align; 3. latent spaces preserve information and are separable in contact shape and object material. In this section, we demonstrate practical use cases for our model via a contact shape reconstruction task. To show that the latent representation can be utilized directly for downstream tasks, we compare two different scenarios: Contact geometry estimation via the latent representation encoded from the same sensor, and encoded from a different sensor.

\subsection{Downstream Data Preparation}

\begin{figure}[b]
    \centering
    \includegraphics[width=1\linewidth]{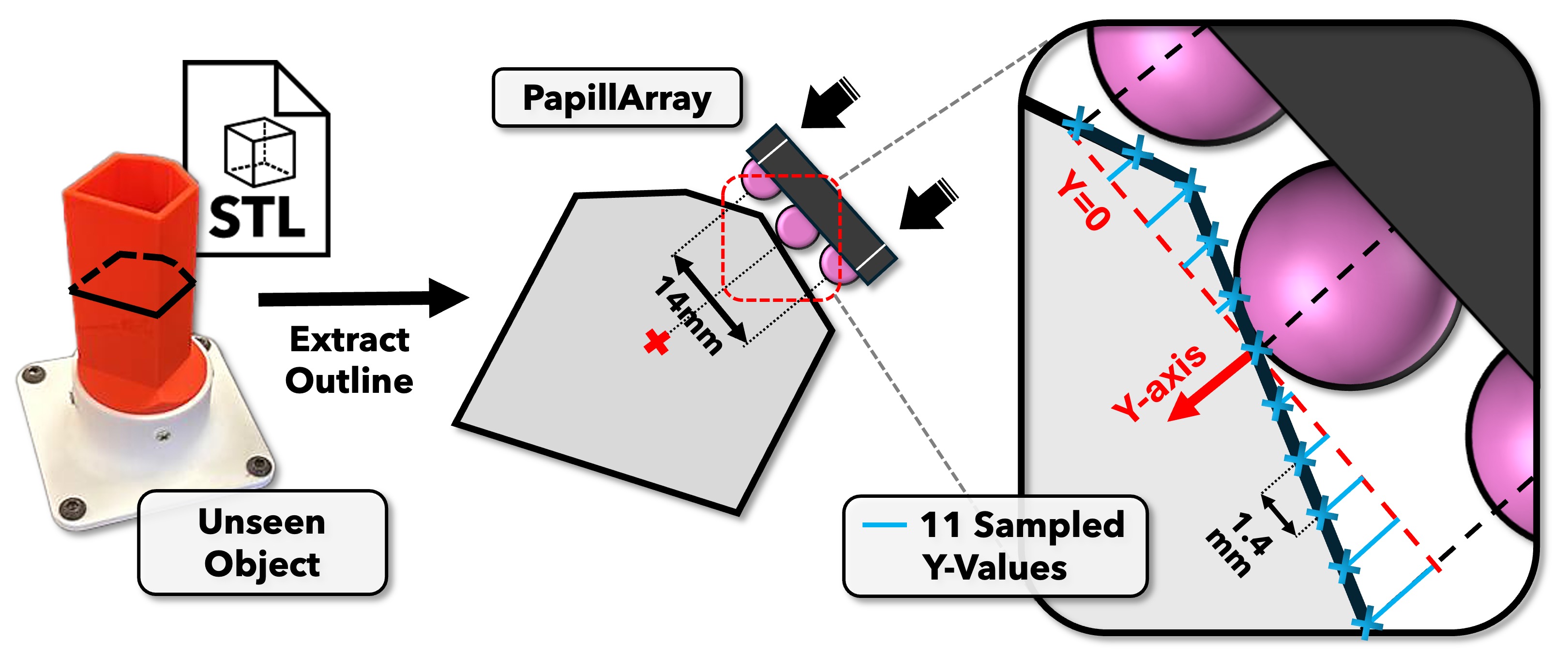}
    \caption{Extraction of contact shape ground truths. A 2D outline from the object's STL file is rotated to align the pressing trajectory with the x-axis, and 11 evenly spaced sampling points are selected along a 14 mm range centered at the contact origin.}
    \label{shape_truth}
\end{figure}

 This task is performed on the unseen object only, which has a more complicated and irregular geometry compared to the seen objects used to train the autoencoder. Following the data-leakage-safe splitting strategy outlined in section \ref{section_dataPrep} and Fig. \ref{trainTest}, we set aside 10\% of the data from the unseen object as the test set (920 samples) and used the remaining 8,280 samples for training the contact shape estimator. 

To extract the contact shape ground truths, the procedure in Fig. \ref{shape_truth} is followed. We extract a high-resolution 2D outline of the unseen object's horizontal cross-section from its STL file. For each approaching angle, a ray is drawn from the object's center (where all pressing trajectories converge), and its intersection with the outline marks the sample's origin. The outline is then rotated around the new origin, aligning the pressing trajectory with the y-axis. Finally, a 14 mm sensing range along the sensing surface (new x-axis) is divided into eleven equal points centered at the origin. 
\subsection{Downstream Model \& Training}
We employ a Multilayer Perceptron (MLP), which consists of four fully connected hidden layers of sizes (64, 128, 64, 32) with dropout layers in between, to map latent representations to an 11-dimensional geometry vector. 80 epochs with a batch size of 64 proved to be sufficient for this task, utilizing the Adam optimizer with a learning rate of 5e-5 and weight decay of 1e-3.

The raw sensor readings are passed through their respective encoders (trained in section \ref{sec3}) to obtain latent space representations. These latent representations are then used to train and evaluate the MLP model. An L1 loss function is used for training, which calculates the average deviation in millimeters across the eleven predicted points and their respective ground truths.

Two models with identical architectures were trained on latent spaces from uSkin or PapillArray and evaluated on two test sets (one from the same sensor and one from the other). Given their differing information content, dropout was set to 0.2 for uSkin and 0.3 for PapillArray.


\subsection{Contact Geometry Estimation Results}

\begin{figure}[!b]
    \centering
    \includegraphics[width=1\linewidth]{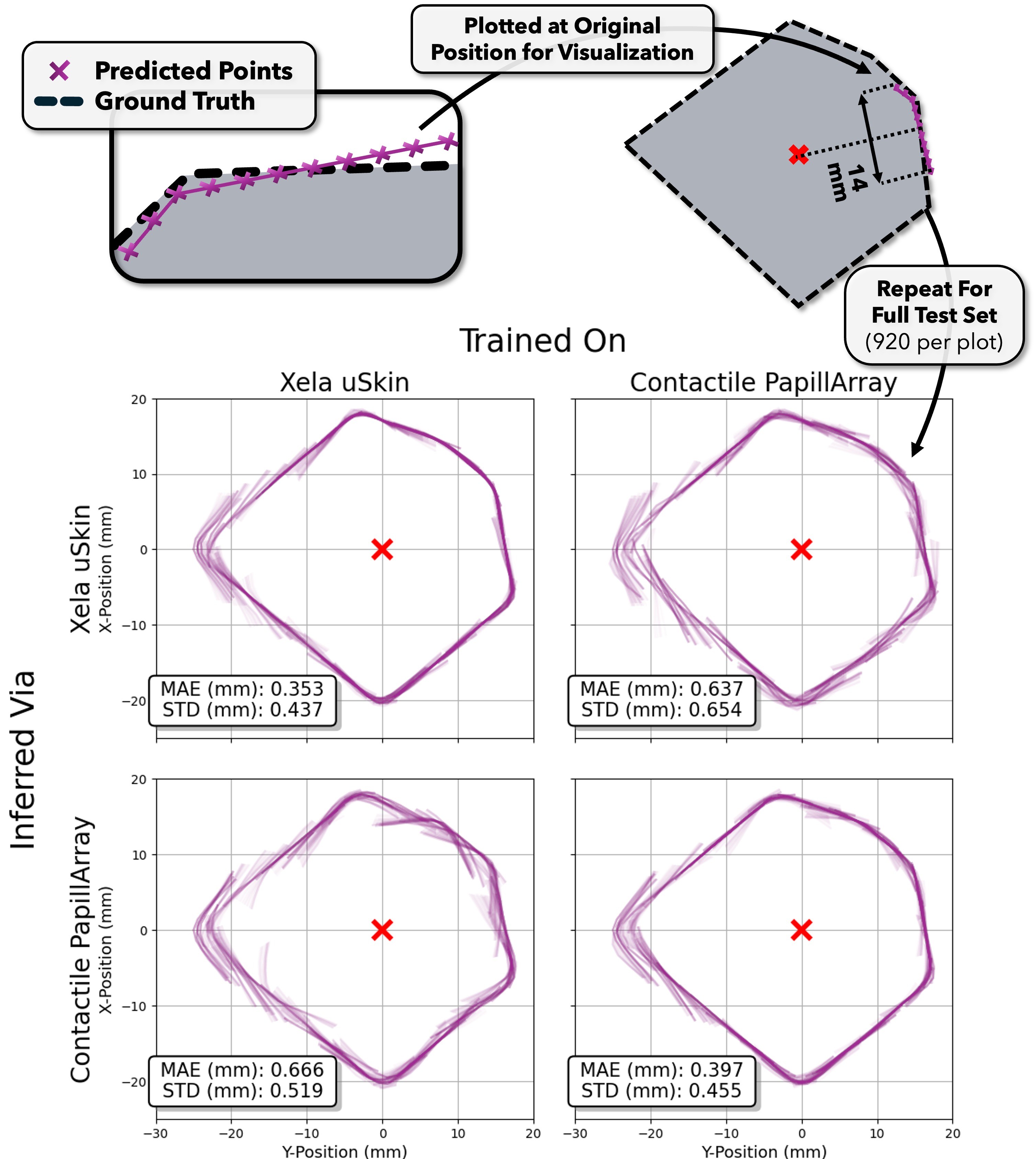}
    \caption{Evaluation of contact shape (local geometry) estimation. Top: Process of mapping predicted local geometries back to their original contact locations for visualization. Bottom: presents all test set predictions across four different scenarios. The columns represent the sensors used for training, and rows indicate the sensors used for testing inference.}
    \label{shape_recon}
\end{figure}


To evaluate the estimated contact shapes, we map the predicted local geometry back to the original contact location on the object for visualization (Fig. \ref{shape_recon}).

All scenarios (single-sensor and cross-sensor) provided satisfactory results, giving clearly recognizable object shapes. Although this irregular object is unseen during UniTac-NV's training, the shared latent space still manages to capture enough information for this task. One noticeable exception is the 80° corner (pointing left on the object), which produces a minimal contact surface. The model correctly detects an edge, but the limited number of activated taxels and sparse shear force data reduce the fidelity of local contact identification, leading to ambiguities.

However, when cross-sensor data transfer is involved (top-right and bottom-left plots in the figure), the contact shape estimations are of a relatively lesser quality. Compared to using the same sensor's unified latent representation for training and inference, the average errors increase from 0.353 mm and 0.397 mm to 0.637 mm and 0.666 mm. 

This performance disparity may be attributed to inherent differences in sensor mechanics. For instance, PapillArray’s design features exposed sensing nubs, with a slightly raised center nub, which enable it to capture shear forces and 3D information more effectively. On the other hand, the uSkin has a larger grid of flat sensing units with a larger contact area. This exposes a limitation for cross-sensor latent space task transfer: only information common between sensors can be transferred from one to another.

\section{Conclusion}
In this paper, we presented a framework for non-vision-based tactile sensor agnostic via an encoder-decoder architecture. By training sensor-specific encoders and a shared decoder on aligned tactile data, our method achieves cross-sensor alignment while preserving information including object material and contact geometric features. The unified latent space representation enables sensor transfer in downstream tasks such as contact geometry estimation, achieving an average estimation error of 0.513 mm. These results confirm the feasibility of cross-sensor generalization but reveal performance variability: When sensor morphologies or modalities differ significantly, only shared information transfers effectively, making hardware capabilities a key factor in task performance.

Our work serves as a proof of concept, demonstrating that tactile data reconstruction can encode transferable physical insights. However, the current dataset, collected on a limited set of sensors and objects with constrained local geometries, restricts generalization to more sensors and unseen contact interactions. We plan on scaling the framework with a larger, more diverse dataset, encompassing more tactile sensors (including vision-based models), broader object shapes, materials, and interaction dynamics, to enhance information preservation and robustness.


\addtolength{\textheight}{-12cm}   




\bibliographystyle{IEEEtran}
\bibliography{references}

\end{document}